\documentclass[twoside,11pt]{article}

\usepackage{blindtext}

\usepackage{jmlr2e}

\usepackage{amsmath}
\usepackage{amsfonts}

\usepackage{lastpage}

\jmlrheading{TBD}{2024}{1-\pageref{LastPage}}{12/25}{12/29}{TBD}{Leo van Nierop}

\ShortHeadings{Transformer models are gauge invariant}{van Nierop}
\firstpageno{1}

\begin{document}

\title{Transformer models are gauge invariant\\A mathematical connection between AI and particle physics}
\author{\name Leo van Nierop \email leo.van.nierop@gmail.com  \\
	\addr Yelp inc. Remote Employee\\
	leo@yelp.com
	}
\editor{None}

\maketitle

\begin{abstract}%
In particle physics, the fundamental forces are subject to symmetries called gauge invariance. It is a redundancy in the mathematical description of any physical system. In this article I will demonstrate that the transformer architecture exhibits the same properties, and show that the default representation of transformers has partially, but not fully removed the gauge invariance.

\end{abstract}

\section{Outline}
This paper consists of two main sections. In the first part I describe the claim of gauge invariance at a high level, and discuss what the consequences of this symmetry are for transformers. I outline practical considerations that will enable reducing the parameters in a transformer based model without any loss in representational power. 

In the second part I go into mathematical detail on how the gauge symmetry can be defined on the transformer architecture, and show that transformer models are indeed invariant: the gauge transformation defines a continuous collection of weights and biasses that all result in the same model function (any given set of inputs leads to the same set of outputs, irrespective of which version of the parameters we are using)

Finally in the conclusion I outline some research directions that this correspondence to gauge theories opens up. 

\section{Introduction}
The transformer architecture was introduced in \cite{vaswani2023attentionneed}, and has been widely used since. For example, the generative LLM revolution of recent years including the gpt family of models (eg. \cite{openai2024gpt4technicalreport}), LLaMA (\cite{touvron2023llamaopenefficientfoundation})  and Gemini (\cite{geminiteam2024geminifamilyhighlycapable}). Those models have very many parameters, and training them is very expensive. This means that if there are inefficiencies in the underlying transformer architecture as I argue in this paper, there is a large opportunity (both financially and in terms of energy consumption). 

\part{Why does this matter}
\section{Parameter reduction}
Deep learning models usually have a function described by very many parameters. Some of those parameters are not relevant to the final model: in the minimum that the model settles in, the value of some matrix elements (or combination of elements) simply has no impact on any of the potential inputs. However, those parameters can still be important in training: it allows the landscape of the loss function to have many directions where the gradient descent can go. This reduces the problem of getting stuck in a local minimum while a significantly better minimum exists elsewhere. However, this only helps when the extra dimension has some gradient in it that leads to a better minimum. If the additional parameter just introduces a dimension which is flat (the loss is the same along the dimension), then it just introduces extra calculation without the benefit of avoiding local minima. The local minimum just turns into a valley rather than a point. 

The symmetry that I show exists in the second part of this paper, is of the latter kind: I identify a set of dimensions that is exactly flat, and the associated parameters are a truly redundant description. Removing them does not remove any paths to a good minimum, but it does reduce the amount of calculation required to find gradients during optimization. It also reduces the amount of operations required at prediction time. When a model is designed in such a way that those redundancies are removed, this represents a cost saving. While we are not talking about orders of magnitude (see table \ref{tab:parameters} for some examples), even a fraction of a percent cost reduction starts to sound relevant with the latest models training cost approaching \$1B\footnote{Maybe not to the companies doing it, but I'll happily pocket a 10\% tip of the savings}.

\begin{table}
	\begin{center}
		\caption{Parameter reduction for common models}
		\label{tab:parameters}
		\begin{tabular}{r|r|r|r}
			\textbf{Model} & \textbf{Parameter count} & \textbf{Redundancy count} & \textbf{Redundancy \%} \\
			\hline
			gpt2 & 117M &1473409 & 1.3 \% \\
			gpt2-XL & 1.56B & 11.1M & 0.7 \% \\
			LLaMA & 65.2B & 201M & 0.3 \% \\
		\end{tabular}
	\end{center}
\end{table}

\subsection{Toy example of a redundant description}
\begin{figure}[h]
    \centering
    \includegraphics[width=\linewidth]{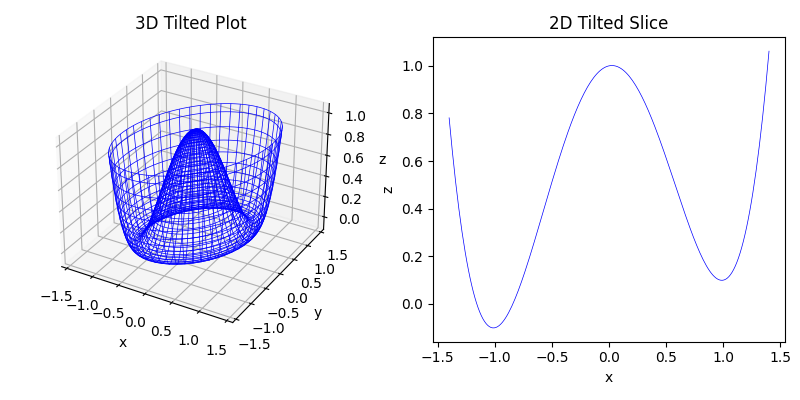} 
    \caption{Example of a helpful new direction avoiding a bad minimum.}
     \label{fig:tilted}
\end{figure}
\begin{figure}[h]
    \centering
    \includegraphics[width=\linewidth]{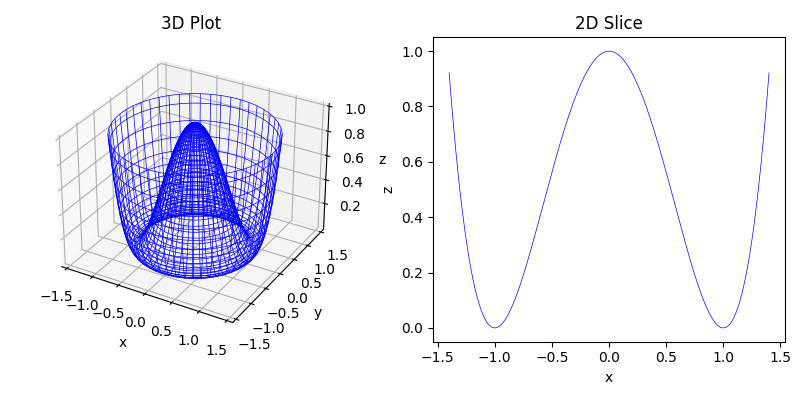} 
    \caption{Example of an unhelpful flat direction.}
    \label{fig:higgs}
\end{figure}

In figure (\ref{fig:tilted}) we see that the 3d plot has a global minimum that can be reached through gradient descent. The 2d plot on the left however has the same global minimum, but it has a second local minimum where the optimization can get stuck. This is an example where the extra parameter allows the model to learn a better minimum, even though in the end the parameter settles at $0$. If we set the parameter to $0$ a priory and remove it from the model, we end up with a worse overall solution (depending on luck with the initialization, of course). Contrast this with the situation in figure (\ref{fig:higgs}). In this situation there is a symmetry, namely rotation around the origin of the $x,y$ plane. The function depends on $x,y$ in a specific way, namely only on the combination $x^2+y^2$ (or, the radius $r$ in polar coordinates).  In this case, setting $y=0$ a priory only influences which of the two identical minima we end up in. This example does not tell us if one of those minima is better in practice, but the same is true in the case of the continuous minimum in the 3d version. In this case, doing the 2d optimization is computationally more efficient without losing out on any paths to the best solution. 

The flat directions I identify in the transformer architecture are of the latter kind: the end to end function that is represented by the transformer architecture is independent of some of the combinations of parameters in the model, forming a higher dimensional sphere rather than a circle like the toy example. 

\part{Why is it true}
\section{Defining the symmetry group}
The symmetries naturally form a group: If a transformation does not change anything about the function we are modeling, then doing another transformation that changes nothing: The combination clearly also changes nothing. This covers closure and associativity. The identity element is clear: don't change anything. Finally, the existence of the inverse of each element. This is not guaranteed in general, but is when we restrict ourselves to mappings that are bijective (one-to-one): we do not include any mappings where two sets of parameters get mapped to the same new set of parameters. 

That being said, there is a natural group that I will pursue in this work. This may not be the full symmetry group of transformer models, but it is the symmetry that I have uncovered with certainty thus far. For an embedding space of dimension $d_e$, the layer normalization restricts the tokens to live on intersection of the unit sphere in $\mathrm{R}^{d_e}$ and the plane through the origin perpendicular to the vector of all ones. This means that they lie on a sphere S($d_e-2$), and the natural invariance of this space is the group SO($d_e-1$), and it will be the subgroup of SO($d_e$) that leaves the vector of all ones invariant. 

In addition, there is a symmetry associated with the dot product attention: Each head of dimension $d_h$ can have the keys and queries transform under opposite representations of $GL(d_h)$ without affecting the attention matrix.

In the next sections, I derive a representation of this group on the parameters of a transformer stack, and show that the model is invariant under those changes. There will be a group element for just the first transformer. I show how to extend the symmetry to have a group element individually for each transformer by extending the transformer architecture to include a rotation in each of the skip connections.  

\subsection{Notation and conventions}
The details of transformers involve a lot of tensor/matrix operations. In order to be able to write them clear and concise, I am adopting the following conventions:
\begin{itemize}
	\item $d_e$ is the number of dimensions of the embedding space. Greek indices from the middle of the alphabet ($\mu,\nu,\rho$ etc.) are used to indicate elements of vectors or matrices in this space.
	\item repeated Greek indices in the same expression are assumed to be summed over (Einstein summation convention), unless otherwise specified.
	\item $n_c$ is the context length: the maximum number of tokens sent into the transformer. In this work I will assume any string send to the transformer is actually exactly of length $n_c$, padded as necessary with a placeholder token.
	\item Latin indices ($i,j,k$ etc.) indicate the tokens. Repeated Latin indices are \emph{not summed} unless explicitly specified
	\item $n_h$ is the number of heads used in the attention mechanism. Latin indices from the start of the alphabet ($a,b$ etc.) indicate the head. Repeated indices are not implicitly summed.
	\item $d_h$ is the dimension of each head (this is often set as $d_h=d_e/n_h$, but this is not necessary so we leave it open here). They are indexed by capital letters from the beginning of the alphabet ($A,B$ etc). When the heads are concatenated, we use $\bar A$ etc. running from 1 to $n_h*d_h$
	\item there are $n_t$ transformers, indexed with greek indices from the beginning of the alphabet  ($\alpha,\beta$ etc.). They are never summed, implicitly or otherwise
	\item The hidden layer in the feed forward network has dimensions $d_f$, and is indexed by capital letters from the middle of the alphabet, ($I,J$ etc.).
\end{itemize}
Finally, I will only use explicit index notation, no dot products etc. If an object has rank $n$, then there should always be $n$ indices on it (or none, if talking about the object in general). If there's indices missing, this is either typo or a math error. 

\subsection{The transformer architecture}
The components of a transformer block as I will consider here are:
\begin{itemize}
	\item layer normalization \begin{itemize}
		\item strict layer normalization (mean subtraction and division by standard deviation)
		\item scale and shift
	\end{itemize}
	\item multi-head attention
	\item skip connections
	\item shared fully connected network with 1 hidden layer
\end{itemize}

\subsubsection{Explicit mathematical formulation of a transformer}
The input to each transformer $T^\alpha_{\mu\nu}$ is denoted $E^{\alpha-1}_{\nu i}$, such that 
\begin{equation}
	E^\alpha_{\mu i} = T^\alpha_{\mu\nu i}E^{\alpha-1}_{\nu i} \quad E^1_{\mu i} = T^1_{\mu\nu i} E^0_{\nu i}
\end{equation}
here $E^0_{\nu i}$ is the initial embedding, and the layer normalized version is $\bar E^0_{\mu i} $. Next, I will break up the transformer, T, into its component steps. The attention matrix is calculated as:
\begin{equation}
A^{\alpha a}_{ij} = \mathrm{Rownorm}\left( \mathrm{Mask}\Bigl( \bar E^{\alpha-1}_{\mu i} Q^{\alpha a}_{\mu A} K^{\alpha a}_{A\sigma} \bar E^{\alpha-1}_{\sigma j} \Bigr) \right)
\end{equation}

Next, the attention is multiplied with the corresponding value matrix and a linear layer:
\begin{equation}
	\hat E^{\alpha-1}_{\bar A i} = \sum_{j} \mathrm{concat}_a \left(A^{\alpha a}_{ij} V^{\alpha a}_{A\mu}   \bar E^{\alpha-1}_{\mu j}  \right) 
\end{equation}

Next there is a connected layer and skip connection:
\begin{equation}
\tilde{E}^{\alpha-1}_{\mu i} =  L^{\alpha-1}_{\mu \bar A} \left(\hat E^{\alpha-1}_{\bar A i} \right)+  E^{\alpha-1}_{\mu i} 
\end{equation}
The row normalization of this is $\bar{\tilde E}^{\alpha-1}_{\mu i}$. Finally the output of the transformer is
\begin{equation}
	E^{\alpha}_{\mu i} = \hat W^\alpha_{\mu I} \mathrm{relu}(W_{I\rho} \bar{\tilde E}^{\alpha-1}_{\rho i}) + \tilde E^{\alpha-1}_{\mu i}
\end{equation}
Relu here is just a sample choice for nonlinearity, the exact function is not relevant to this derivation. 

\subsection{What is a gauge transformation for a transformer}
A trained transformer consists of a variety of weight matrices (Q, K, V,  $\hat W$, W, L). To apply the transformer, a set of $n_c$ token embeddings gets mapped to a new set of $n_c$ token embeddings. To apply a gauge transformation to the transformer means to change the weight matrices according to a specific rule. The rule needs to be specified without knowing what the weight matrices will be, so it could be applied at any time before, during or after training. Furthermore, the transformation needs to be invertible, so it forms a group. 
Finally, it is allowed to also apply a specific group element to the token embeddings at the start of the transformer stack, and at the end (ie. just after going from strings to vectors, and just before going from vectors to probability distributions over predicted next tokens). Those beginning and end transformations can be thought of as modifying the encoding and decoding matrices, but since we are looking at just the transformers here we allow them to apply on the embeddings directly.

\subsubsection{What does invariance mean for a transformer}
A transformer based self-attention model is invariant under a gauge transformation, if for all possible token inputs, the predicted probability distribution for the following tokens is identical before and after the gauge transformation. The intermediate states in the model can, and typically will be different, but the final output is unchanged (in practise, that would be unchanged up to machine precision).

\subsection{Derivation of the invariance choices}
In the following, I am denoting the group $G=SO(n_e-1)$, and any $g^{(i)}_{\mu\nu}$ are elements of the representation of this group in $\mathbb R^{n_e}$ that leaves the vector of all ones invariant. Furthermore I will denote the group $H=GL(d_h)$, and  $h^{(i)}_{AB}$ are  elements of the group acting on the heads. Similarly $F=GL(d_f)$. We consider the following transformations on the parameters of the transformer, and find the constraints required for invariance. I do this on the first transformer in the stack, but this generalizes trivially:
\begin{align}
	E^0_{\mu i} &\rightarrow g^{(0)}_{\mu\nu} E^0_{\nu i} \\
	\bar E^0_{\mu i} &\rightarrow g^{(0)}_{\mu\nu} \bar E^0_{\nu i} \\
	K^{0a}_{A\mu} &\rightarrow h^{(1a)}_{AB} K^{0a}_{B\nu} g^{(1a)}_{\mu\nu}\\
	Q^{0a}_{A\mu} &\rightarrow h^{(2a)}_{AB} Q^{0a}_{B\nu} g^{(2a)}_{\mu\nu}\\
	V^{0a}_{A\mu} &\rightarrow h^{(3a)}_{AB} V^{0a}_{B\nu} g^{(3a)}_{\mu\nu}\\
	L^{0}_{\mu \bar A} &\rightarrow g^{(4)}_{\mu\nu} L^{0}_{\nu B}  \bar h^{(4)}_{\bar A\bar B} \\
	W^0_{I\rho} &\rightarrow  f^{(0)}_{IJ}W^0_{J\mu}g^{(5)}_{\mu\rho} \\
	\hat W^0_{\mu I} &\rightarrow g^{(6)}_{\mu\nu} \hat W^0_{\nu J} f^{(1)}_{IJ}
\end{align}
The condition that the attention matrix is invariant becomes:
\begin{align}
	g^{(1a)\ T} = g^{(2a)\ T} = g^{(0)\ -1} \\
	h^{(2a)\ T} = h^{(1a)\ -1}
\end{align}
The first condition just tells us that the first three elements of $G$ that we use are identical. From applying the values matrix similarly we also have $g^{(3)}=g^{(0)}$. The second condition means that we have 1 choice in $H$, and the other element is then fixed. From the linear layer we get the condition
\begin{equation}
	\bar h^{(4)}_{\bar A\bar B} \cdot \mathrm{diag}_a(h^{(3a)}_{\bar B \bar C}) = 1
\end{equation}
The skip connection forces $g^{(4)}=g^{(5)}=g^{(0)}$, and the nonlinearity in the feed forward network forces $f^{0}=f^{1}=1$. The final skip connection gives $g^{(6)}=g^{(0)}$. 

Combined, this leaves us with a single choice of $g$, and two choices of $h^a$. The choice of $g$ also affects the output of the transformer, so it can only be chosen once at the first transformer, and then must be the same for the subsequent ones. However, the choice of $h$ is fully internal to the transformer so can be chosen independently on each transformer. 

\subsection{calculation of the number of redundant dimensions}
From the calculation above, we see that we have the choice of a single element of $SO(n_e-1)$, and $2*n_t*n_h$ choices of an element of $GL(d_h)$. That means the total redundancy in a transformer stack is
\begin{equation}
	\text{Redundancy}=2n_tn_hd_h^2 + \frac12(d_e-1)(d_e-2)
\end{equation}
For the evaluation of this redundancy on some well known models, see table \ref{tab:parameters}

\subsection{Enlarging the symmetry group}
The symmery of the embedding space, $(n_e-1)$, is only applied once. The reason that it isn't applied  on each transformer individually, is because of the skip connections. They are forcing the embedding space to align at different positions in the transformer stack. This is essentially a choice of gauge, and we can undo that choice enlarging the symmetry group. In order to do so we need to introduce additional parameters like this:
\begin{align}
	\tilde E^{\alpha-1}_{\mu i} &= L^{\alpha-1}_{\mu \bar A}\hat E^{\alpha-1}_{\bar A i} + G^{\alpha-1}_{\mu\nu}E^{\alpha-1}_{\nu i} \\
	E^\alpha_{\mu i} &= \hat W^{\alpha -1}_{\mu I}\mathrm{relu}(W_{I\rho}\bar E^{\alpha-1}_{\rho i}) + \bar G^{\alpha-1}_{\mu\nu} \tilde E^{\alpha-1}_{\nu i}
\end{align}
They transform under the group $G$,
\begin{align}
	G^{\alpha-1}_{\mu\nu} &\rightarrow g^{(7)}_{\mu\rho}G^{\alpha-1}_{\rho\sigma}g^{(8)}_{\nu\sigma} \\
	\bar G^{\alpha-1}_{\mu\nu} &\rightarrow g^{(9)}_{\mu\rho}\bar G^{\alpha-1}_{\rho\sigma}g^{(10)}_{\nu\sigma} \\
\end{align}
The conditions for invariance change such that $g^{(4)}=g^{(7)}=g^{(10)}$, $g^{(8)}=g^{(0)}$, $g^{(9)}=g^{(6)}$, and the final output of each transformer goes to $E^\alpha_{\mu i} \rightarrow g^{\alpha-1 (6)}E^\alpha_{\mu i}$, so $g^{\alpha-1(6)}=g^{\alpha (0)}$, and on each transformer we have the freedom to pick an individual $g^{(0)}$ and $g^{(4)}$.

This does not change anything about the number of flat directions identified, since we introduce two parameters that are elements of $SO(n_e-1)$ and two gauge choices in the same group. The net number of parameters in the model remains as before. This shows that the regular transformer architecture is a gauge-fixed, discretized representation of a $SO(n_e-1)$ gauge field theory.

\section{Related work}
In \cite{loshchilov2024ngptnormalizedtransformerrepresentation} the authors redefine the transformer architecture to have all weights and representations to live on the hypersphere. The hypersphere as described here is what in a gauge theory representation is the fundamental representation of the gauge group. Gauge transformations are rotations of that sphere, basically choosing a new coordinate system between each set of transformers. The gauge theory representation can be used in their work by representing the weights matrices etc. as elements of the Lie algebra of $SO(n)$, maintaining the normalization of the various vectors automatically rather than having to enforce it after each training step. 

\section{Conclusions}\label{conclusions}
I have shown that a stack of transformers can be transformed (the weights changed according to specific rules) without affecting the final output of the model, regardless of what the input tokens are. This means that the transformed weights model the exact same function as the original weights. The group elements that define the transformations give rise to exactly flat valleys in the loss function. The flat directions represent parameters for which calculations are done that are not necessary, both during training and during inference. The gauge transformations derived in this paper can be used to specify a gauge choice, where some of the parameters in the transformer model can be removed by replacing them with the identity operator, passing through any calculations. Additionally, the connection between transformers and gauge theories opens avenues of research on understanding why transformers work, and how they can be represented. 

\subsection{Thoughts on future research potential}
\subsubsection{Gauge theory properties are well researched}
Gauge theories have been extensively studied in physics starting with \cite{PhysRev.96.191}, becoming quite a mature subject in the late 70's and 80's. Given their prominence in theoretical physics, their general properties are well known now. Knowing the extend to which the transformer architecture and gauge field theory map to one another allows us to apply this prior research to transformer models. One example of gauge theory properties that could be relevant is the existence of topological constraints. Many gauge theories contain classes of solutions that cannot continuously be transformed into one another. The equivalence for transformers is likely that the seed weights can be very important: If the seeds are in a topological class that is different from the topological class of the solution that best fits the data. In that case, gradient descent cannot reach the optimal solution. It may be worthwhile to train smaller models with the same topological properties to determine the right topological class, and then seed a large model constraint to lie in that class. 

\subsubsection{Alternative representations of the same capabilities}
A regular transformer stack can be seen as a discretized version of time evolution of a gauge field. The time direction in this case is along the direction of subsequent transformers, while the context (number of tokens) can be thought of as the spatial direction in the model\footnote{It's also possible to flip this, with the benefit that the time ordering coincides with the token ordering, so that the future mask is the same as causality in the gauge theory. This is outside of the scope of the current paper.}. The transformer's skip connections mean that the transformers are mappings of a compact space to itself. The same mapping can probably also be described as the solutions to a differential equation, where the initial tokens live at time $t=0$, while the outputs live at $t=1$. The transformer stack then is a discretization of the differential equation, with a deeper stack representing a smaller time step (closer approximation to the continuous solution). Alternatively, the time direction could be expanded in a Fourier series, where subsequent terms in the series give a better approximation of the highly variable language content. Such an approach allows the depth of the transformer stack to be unspecified at first, expanding the series until the desired accuracy has been reached. At this point I do not know if such an approach carries practical benefits, but it opens up an area of research into the properties of transformers that I believe is worth pursuing. 

\vskip 0.2in
\bibliography{transformers_jmlr}

\end{document}